\documentclass{article}

\PassOptionsToPackage{numbers, compress}{natbib}

\usepackage[final]{neurips_2025}

\usepackage[utf8]{inputenc}
\usepackage[T1]{fontenc}
\usepackage{hyperref}
\usepackage{url}
\usepackage{booktabs}
\usepackage{amsfonts}
\usepackage{nicefrac}
\usepackage{microtype}
\usepackage{xcolor}
\usepackage{graphicx}
\usepackage{amsmath,amssymb}
\usepackage{multirow}
\usepackage{array}
\usepackage{enumitem}
\usepackage{algorithm}
\usepackage{algpseudocode}
\usepackage{tikz}
\usetikzlibrary{arrows.meta,positioning,fit,shapes.geometric}

\usepackage{xpatch}
\makeatletter
\xapptocmd{\NAT@bibsetnum}{\setlength{\leftmargin}{0pt}\setlength{\itemindent}{\labelwidth}\addtolength{\itemindent}{\labelsep}}{}{}
\renewcommand{\@notice}{}

\makeatother

\newcommand{\sys}{LemonHarness}
\newcommand{\tb}{Terminal-Bench}

\title{\sys{} Technical Report}

\author{
Kailong Ren$^{1,*}$, Fubo Sun$^{1,*}$, Jiachen Liu$^{1}$, 
Liu Yang$^{1}$, Zimo Yin$^{1}$, Jiaying Li$^{1}$,\\
\bfseries Congli Yin$^{1}$, Ming He$^{1,\dagger}$, Yu Huo$^{1}$,
Jiawei Liu$^{1}$, Zeping Chen$^{1}$,\\
\bfseries  Yubin Huangfu$^{1}$,
Ronghua Li$^{1}$,Yixuan Wu$^{1}$, Xing Su$^{1}$, Yanzhi Xu$^{1}$,\\
\bfseries 
Likang Wu$^{2}$, Hongke Zhao$^{2}$, 
Lei Zhang$^{3}$,
Xiaohui Geng$^{4}$,
Jianping Fan$^{1}$\\[1ex]
$^{1}$AI Lab @ Lenovo CTO Org\\
$^{2}$Tianjin University\\
$^{3}$Anhui University\\
$^{4}$Department of Psychological and Cognitive Sciences, Tsinghua University
}

\begin{document}

\maketitle

\begingroup
\renewcommand{\thefootnote}{}
\footnotetext{%
\parbox[t]{0.9\textwidth}{%
$^{*}$Equal contribution.\\
$^{\dagger}$Corresponding Author. Email: \texttt{heming01@foxmail.com}
}%
}
\endgroup
    
\begin{abstract}
As large language model (LLM) agents are applied to longer  tasks, they increasingly modify workspace state across multiple rounds of iteration. However, agents typically observe only tool outputs and log fragments, while the actual state changes occur in the file system. Without explicit workspace boundaries, state-changing operations such as file writes and temporary artifact generation may scatter changes across paths. Over time, these weakly constrained changes accumulate, making states such as modified files difficult to track.
This paper presents \textbf{LemonHarness}, an integrated execution framework for long-horizon agents. LemonHarness establishes an explicit execution boundary by constraining state-changing operations within a clearly defined workspace and bringing model invocation, tool execution, and rule knowledge within a single controlled boundary. State-changing operations, including file writes, dependency installation, and temporary artifact creation, are executed through structured tool interfaces, with execution feedback recorded as observations available to subsequent model decisions.
The system also introduces a reusable rule knowledge base, which turns recurring execution rules and acceptance criteria into runtime knowledge. LemonHarness further adds a time-aware execution mechanism that exposes elapsed and remaining budget to the model, so it can rebalance exploration, implementation, and validation effort as time pressure shifts and avoid timeouts from long waits or excessive verification.
On Terminal-Bench 2.0, LemonHarness\_GPT-5.3-CodeX reached 84.49\% accuracy over 445 trials; pairing the same framework with the stronger GPT-5.5 backbone raised the average accuracy to 86.52\% across five jobs. The results suggest that a unified runtime boundary, callable rule knowledge, and time-aware execution can improve the stability of long-horizon agent execution.
\end{abstract}

\section{Introduction}
LLM agents are increasingly used for tasks that are longer, more stateful, and closer to real execution environments~\cite{erdogan2025planandact,luo2025ultrahorizon,song2026envscaler,lin2026agentic}. In these tasks, a model may need to call tools, edit files, install dependencies, create temporary artifacts, start external processes, and revise later actions based on intermediate results. 
The key challenge is therefore not only whether each individual reasoning step is correct, but also whether the agent can maintain an accurate view of the workspace after many state-changing operations.
In practice, the model usually observes only tool outputs, error messages, and log fragments, while the actual state changes occur in the shell, file system, temporary directories, dependency environment, and external processes. 
Without a clear workspace boundary and explicit constraints on state-changing actions, file writes, dependency installation, and temporary directory creation may be scattered across different paths and processes.
As the execution chain grows longer, these weakly constrained state changes accumulate, making it difficult for the model to determine which files have been modified, which commands have changed the environment, and which validation results remain valid, thereby causing state drift.

SABER~\cite{cuadron2025saber} shows that failures in long-horizon agents are not spread evenly across all steps. They concentrate around a small number of mutating actions, such as file creation. These actions account for only about 14--18\% of all steps, but one incorrect mutating deviation can reduce task-success odds by 57--82\%, while a similar non-mutating deviation reduces success odds by only 7--15\%. Other studies also suggest that long-horizon task success depends not only on individual actions, but also on whether the environment state remains consistent across multiple rounds of execution~\cite{shen2026evocode,wang2026long}. 
These findings suggest that state-changing operations in long-horizon tasks should be executed under explicit workspace constraints, which can mitigate state drift.

Motivated by this observation, LemonHarness makes the execution boundary explicit. It constrains state-changing operations within a clearly defined workspace and organizes model invocation, tool execution, structured tool definitions, task knowledge, workspace state, and execution logs within the same controlled framework.
State-changing operations such as file writes, dependency installation, temporary artifact creation, and validation commands are executed through structured tool interfaces inside a designated workspace. Their feedback is recorded as observations for later decisions, so the model can continue from a coherent workspace state rather than infer the environment from scattered logs. This design keeps observation, execution, modification, and verification aligned around the same task context, reducing the risk of state drift in long-running tasks.

Prior studies have shown that reusable experiential knowledge can effectively improve model performance and execution stability~\cite{shinn2023reflexion,wang2026webxskill, yang2026skillopt}. LemonHarness further introduces general procedural knowledge by transforming task-transferable execution heuristics and validation criteria into runtime priors. Many tasks share reusable execution patterns. For example, in deep learning tasks, a decreasing loss alone does not necessarily indicate task completion; general constraints such as data splits, random seeds, and validation metrics must also be considered. If these constraints are not incorporated into runtime knowledge, the model may produce solutions that appear plausible but are invalid in practice. General procedural knowledge exposes such task boundaries in advance, giving the agent an early sense of direction and thereby enabling more stable task execution.

Long-horizon tasks are constrained not only by correctness requirements but also by strict execution budgets~\cite{kamath2025enforcing, paglieri2025learning}. In terminal environments, an agent may spend too much time on open-ended exploration, wait for long-running commands, or repeatedly verify intermediate results even when an acceptable solution has already been produced. Such behavior can turn an otherwise solvable task into a timeout failure. To address this issue, LemonHarness introduces a time-aware execution mechanism. Before execution, LemonHarness assigns the task to a time tier and then feeds elapsed time and remaining time back to the model during execution. With this signal, the model can shift effort between exploration, implementation, and validation as the deadline approaches, instead of stalling on long-running commands or piling on redundant late-stage checks.

On Terminal-Bench 2.0~\cite{merrill2026terminal}, LemonHarness\_GPT-5.3-CodeX achieved 84.5\% accuracy over 445 trials. In additional validation with GPT-5.5~\cite{singh2025openai}, LemonHarness reached 86.52\% average accuracy across five jobs, solving 385 out of 445 trials. These results suggest that the combination of a unified runtime boundary, reusable general rule knowledge, and time-aware execution can deliver strong performance on realistic long-horizon tasks.

The main contributions are as follows:
\begin{itemize}[leftmargin=*]
    \item LemonHarness designs an integrated execution framework that manages prompts, tool executors, skills, execution logs, and model calls in one place, keeping observation, execution, modification, and verification under the same runtime boundary.
    \item LemonHarness introduces reusable general rule knowledge, turning recurring execution rules and acceptance criteria into runtime priors that guide the model early in a task.
    \item LemonHarness adds a time-aware execution mechanism, allowing the model to adjust exploration, implementation, and validation according to elapsed and remaining time.
\item LemonHarness is evaluated on Terminal-Bench 2.0, where LemonHarness\_GPT-5.3-CodeX achieved 84.5\% accuracy over 445 trials, and additional validation with GPT-5.5 reached 86.52\% average accuracy across five jobs.
\end{itemize}

\section{Related Work}

\subsection{Integrated LLM Agent Frameworks}
Recent work has explored how to integrate reasoning, acting, and tool usage within unified agent frameworks~\cite{ning2026code, lin2026agentic, xu2026adapting}. 
ReAct~\citep{yao2022react} interleaves chain-of-thought reasoning with tool invocation, demonstrating that reasoning traces can guide action selection in a single execution loop. 
Toolformer~\citep{schick2023toolformer} enables language models to self-supervise API usage by inserting tool calls into training data, providing a mechanism for structured external interaction. 
PAL~\citep{gao2023pal} separates natural language reasoning from program execution by delegating computation to external interpreters, emphasizing modular execution. 
Voyager~\citep{wang2023voyager} further introduces a lifelong skill library that accumulates executable programs over time, forming a persistent action space. 
While these approaches connect models with tools or execution environments, they typically treat prompting, execution logs, and runtime control as loosely coupled components. 
In contrast, LemonHarness unifies prompts, tool executors, skill management, execution traces, and model calls within a single runtime boundary, enabling consistent observation, modification, and verification under one integrated execution framework.

\subsection{Reusable Knowledge and Time-Aware Control}

Several recent agent systems reuse knowledge from previous experience or discovered action patterns. ExpeL extracts natural-language insights from past trajectories, while WebXSkill converts recurring web-action subsequences into reusable skills~\cite{zhao2024expel,wang2026webxskill}. LemonHarness follows this direction but focuses on general execution rules. It turns recurring constraints, evaluator requirements, and acceptance criteria into runtime priors, so the agent can begin a task with clearer boundaries before detailed exploration.

Beyond knowledge reuse, long-horizon agents face a second orthogonal challenge: managing time. Long-horizon agents must also decide how much effort to spend on exploration, implementation, and validation. Prior work on dynamic planning shows that planning before every action can introduce unnecessary test-time cost in long-horizon settings~\cite{paglieri2025learning}. LemonHarness makes this tradeoff explicit through time-aware execution: elapsed and remaining time are treated as runtime state, allowing the agent to shift from broad exploration toward result preservation and targeted validation as the deadline approaches.
\section{Method}

\subsection{Overall Architecture}

\begin{figure*}[t]
    \centering
    \includegraphics[width=0.85\textwidth]{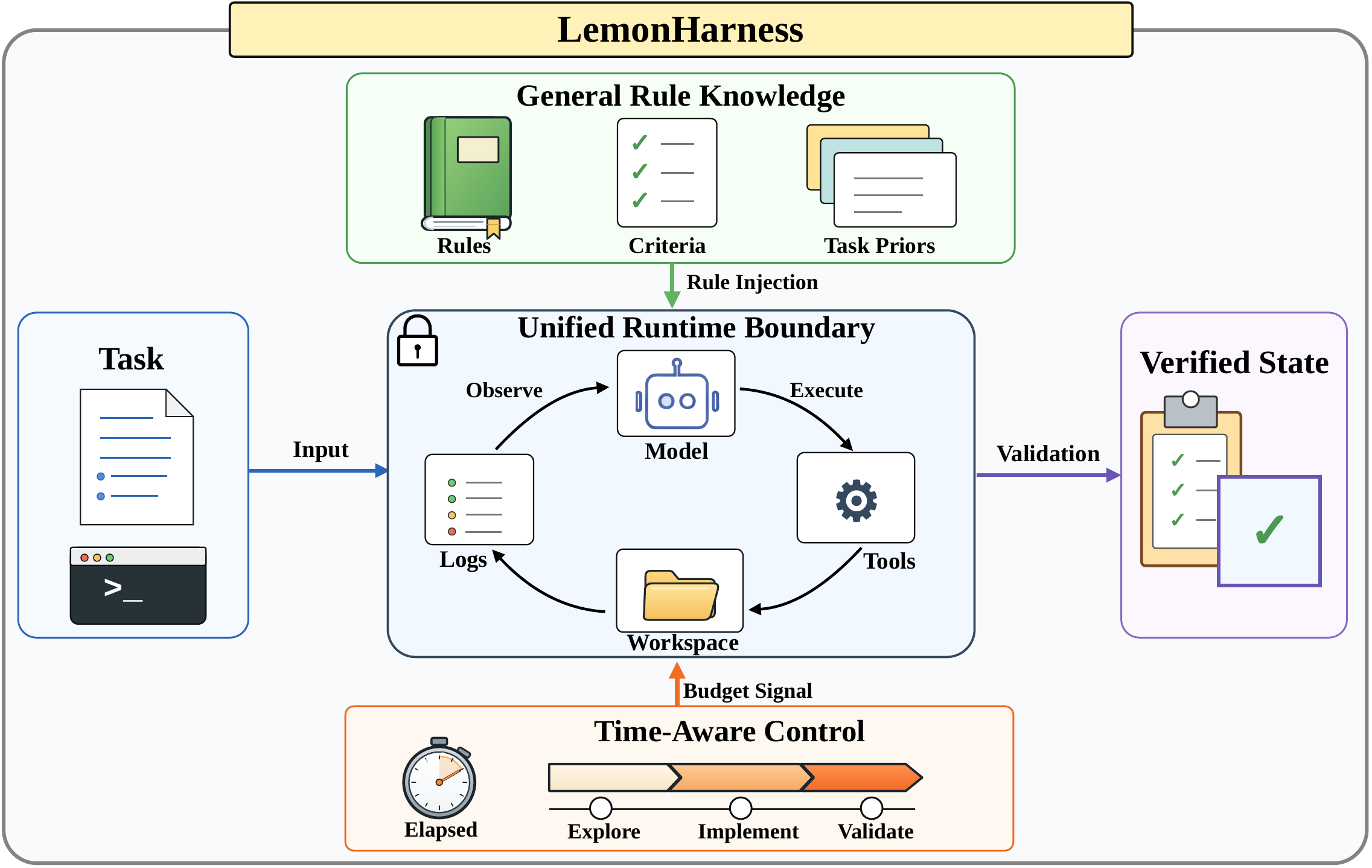}
    \caption{Overview of the LemonHarness framework. A task is processed through a unified runtime boundary that integrates model reasoning, tool execution, workspace state, and execution logs. General rule knowledge provides reusable execution rules, criteria, and task priors, while time-aware control adjusts exploration, implementation, and validation according to elapsed and remaining budget. The final task state is checked through external validation, while traceable execution records are retained.}
    \label{fig:framework}
\end{figure*}

LemonHarness builds a layered execution framework for long-horizon tasks. The system is organized around task intake, model decision-making, tool execution, knowledge support, and log recording. When a task enters the system, the runtime first creates a controlled workspace and loads the system prompt, tool definitions, and generalizable rule knowledge. The model then produces an action intention based on the current task state; this intention is translated into a structured tool call, executed inside the controlled workspace, and returned as feedback for later decisions. Figure~\ref{fig:framework} gives an overview of this execution framework.
By scheduling these steps within the same execution boundary, LemonHarness keeps model judgments, tool actions, and environment state aligned around a shared task context. A time-aware mechanism feeds remaining budget back to the model, so it can pace inspection, building, and validation according to how much time is left, rather than stalling on long-running commands or accumulating redundant checks near the deadline.

These design choices allow the agent to keep environment state and time pressure synchronized across long execution chains, while generalizable rule knowledge helps it establish a feasible direction early. Together, they make task execution more stable.

\subsection{Integrated Execution Framework}

The core method of LemonHarness is to bring the main components of agent execution into one framework. The system prompt, tool executor, skill library, execution logs, and model-control loop are managed together rather than being left as separate pieces around the model. This design puts the agent's observation, action, modification, and verification steps under the same set of runtime constraints.

For long-horizon tasks, this integrated boundary is important because the model's belief about the task can easily drift away from the actual environment state. A file may have been modified, a dependency may have been installed, or a command may have changed the workspace in ways that affect later decisions. By keeping model control, tool execution, task knowledge, and logs inside the same framework, LemonHarness reduces this disconnect and makes the execution process easier to track and correct.

\subsection{Reusable Rule Knowledge}

LemonHarness organizes reusable execution rules and acceptance criteria as runtime knowledge, so the model does not have to start each task from a blank slate. This knowledge captures patterns that recur across task families, including input-output conventions, evaluator constraints, environment restrictions, and common failure modes. It gives the agent an early sense of what a valid solution should look like before it begins detailed exploration.
The role of generalizable rule knowledge is to expose these boundaries up front, thereby reducing blind trial and error and aligning the agent's early actions with the acceptance conditions the evaluator will ultimately apply.

Table~\ref{tab:rule-skills} summarizes the current coverage of reusable rule knowledge in LemonHarness. Across these areas, the knowledge base captures recurring input-output contracts, environment assumptions,  and validation orders.

\begin{table*}[h]
\centering
\caption{Coverage of reusable rule knowledge in LemonHarness. Each area is supported by reusable rules that capture recurring constraints and validation habits for a task family.}
\label{tab:rule-skills}
\renewcommand{\arraystretch}{1.12}
\small
\begin{tabular}{@{}>{\raggedright\arraybackslash}p{0.24\textwidth}>{\raggedright\arraybackslash}p{0.70\textwidth}@{}}
\toprule
Coverage area & Rule Knowledge Overview  \\
\midrule
ML workflows & Keep training, validation, and submission artifacts separate; preserve reproducibility; validate metrics against the task goal. \\
Bio-design & Retrieve source data through stable APIs; preserve biological validity; check final constructs against synthesis constraints. \\
Vision media & Keep command interfaces stable; preserve row, frame, or mask alignment; validate serialized visual outputs before submission. \\
Systems recovery & Work inside controlled paths; create recoverable artifacts early; use build, integrity, and runtime probes to avoid blind repair. \\
Game logic & Preserve formal rules and state transitions; bound strategic search; verify the final state against scoring conditions. \\
\bottomrule
\end{tabular}
\end{table*}
\subsection{Time-Aware Execution}

LemonHarness introduces a time-aware execution mechanism that treats time as part of the runtime state rather than as an external timeout only.  At the start of
a task, the system assigns a total time budget~$T$ according to the
task's expected difficulty and execution cost. During execution, elapsed
time and remaining time are surfaced to the model at every turn, so the
agent can judge not only what has been done, but also how much time is
left to finish.

\paragraph{Phase schedule.}
Within~$T$, \sys{} partitions execution into three behavioral phases
using cumulative budget ratios, mirroring the explore--implement--validate
rhythm that long-horizon problem solving naturally follows. Each transition
is guarded by a small grace band, so the agent is not forced to abandon a
near-finished step on the exact boundary. Defaults are summarized in
Table~\ref{tab:phase}.

When elapsed time crosses a phase boundary (e.g.,~$0.3T$), the runtime
does not switch immediately; instead it grants a $0.05T$ grace window
before instructing the model to advance to the next phase. This avoids
cutting off an exploration step that is one or two commands from
completing, while still preventing open-ended drift. Transitions are
monotonic: once the agent enters P2 or P3, it does not roll back to an
earlier phase even if a later step turns out to be cheap, which keeps
the schedule stable against short-term timing noise. The final $0.1T$
(P4) is treated as a hard reserve in which the model is instructed to
stop initiating new state-changing actions and preserve whatever
acceptable result is already on disk.

Across all four phases, the model receives an explicit signal at every
turn indicating both the current phase label and the remaining fraction
of~$T$. This converts timing from an opaque external deadline into a
first-class input to the decision loop. The phase schedule does not
prescribe specific actions within each window; rather, it tilts the
agent's prior so that early turns favour broad inspection while later
turns favour preserving a working result over pursuing a marginally
better one. The net effect is that timeouts from runaway exploration or
last-minute over-checking are avoided not by hard interruption, but by
a continuous adjustment of the agent's own priorities.
\begin{table}[h]
\centering

\caption{Default phase schedule used by time-aware execution. Ratios are
cumulative fractions of the total task budget~$T$. The grace band allows
the agent to finish an in-flight step that is close to completion before
being prompted to advance.}
\label{tab:phase}
\small
\resizebox{\linewidth}{!}{%
\begin{tabular}{llllc}
\toprule
Phase & Budget window & Parameter (default) & Strategy & Grace \\
\midrule
P1 Explore   & 0\%--30\%   & \texttt{explore\_ratio}~0.3   & Inspection, planning, environment setup        & 5\% \\
P2 Implement & 30\%--60\%  & \texttt{implement\_ratio}~0.6 & Primary solution construction                  & 5\% \\
P3 Validate  & 60\%--90\%  & \texttt{validate\_ratio}~0.9  & Lock in result, targeted verification          & 5\% \\
P4 Reserve   & 90\%--100\% & --                            & Preserve output, no new state-changing actions & --  \\
\bottomrule
\end{tabular}%
}
\end{table}

\subsection{Structured Tool Boundary}

LemonHarness provides environment operations through structured tools. Capabilities such as file reading, file editing, command execution, and web access are wrapped as explicit interfaces with defined inputs, outputs, and usage constraints. The model observes and modifies the environment through these interfaces, which makes each operational step easier to interpret and record.

State-changing actions receive stricter treatment. File writes, temporary directory creation, dependency installation, and background process startup can all affect later execution. LemonHarness keeps these actions inside the controlled workspace and the structured tool-call boundary, so their effects remain tied to the current task context. This reduces the chance that the meaning of an action is lost or misread in a long context, and it helps the model reason later about which parts of the environment have changed.

\subsection{Execution Records and Validation Feedback}

Long-horizon tasks require intermediate checks throughout execution, rather than leaving all risk to the final step. LemonHarness records model outputs, tool calls, and execution feedback in runtime logs, and feeds tool results back into the later context. This allows the model to revise its plan based on the current state instead of restarting from scattered observations or repeating the same exploration.

Validation feedback is treated as part of the execution process. Commands used for testing, verifier-like checks, file modification results, and error messages are kept as traceable records. These records help the agent understand what has already been tried, which changes affected the environment, and where a failure may have entered the task trajectory. They also provide evidence for later failure analysis and targeted correction.

\section{Experiments}

\subsection{Experimental Setup}

LemonHarness is evaluated on Terminal-Bench, which covers a range of real terminal-task scenarios. Each trial asks the agent to follow a natural-language instruction, operate on the environment, modify files, and reach a final state accepted by the evaluator.

For the Terminal-Bench 2.0 submission, the configuration name is LemonHarness\_GPT-5.3-CodeX. The main model is GPT-5.3-CodeX, with gemini-3.1-pro-preview used as an auxiliary model for image and video question answering.

\subsection{Terminal-Bench 2.0 Results}

LemonHarness\_GPT-5.3-CodeX submitted 5 jobs on Terminal-Bench 2.0, covering 445 trials in total. Each job contained 89 trials. The system solved 376 trials, failed 67 trials, and had 2 trials without reward. 
As shown in Table~\ref{tab:jobs}, the accuracy of the five jobs ranged from 80.90\% to 89.89\%, with an average accuracy of 84.49\%. On average, each job solved 75.2 trials. The roughly 9-point spread across jobs suggests that long-horizon agent evaluation is still affected by model sampling, task-path choices, and model-interface stability.

The remaining failures are mainly concentrated in tasks involving long compilation, training, multimodal processing, and cross-environment adaptation. These tasks usually require longer execution chains and stricter intermediate validation. Across 445 trials, 39 trials reported exceptions, mostly due to task timeout. This indicates that future work should further improve long-task planning and control of unproductive exploration.
\begin{table*}[h]
\centering
\caption{\tb{} 2.0 results across five validated jobs.}
\label{tab:jobs}
\begin{tabular}{lccccc}
\toprule
Job date & Trials & Solved & Failed & Exception & Accuracy \\
\midrule
2026-04-25 & 89 & 76 & 12 & 7 & 85.39\% \\
2026-04-26 & 89 & 80 & 9 & 4 & 89.89\% \\
2026-04-27 & 89 & 72 & 17 & 14 & 80.90\% \\
2026-04-28 & 89 & 72 & 16 & 7 & 80.90\% \\
2026-04-29 & 89 & 76 & 13 & 7 & 85.39\% \\
\midrule
Average & 89 & 75.2 & 13.4 & 7.8 & 84.49\% \\
\bottomrule
\end{tabular}
\vspace{-0.1in}
\end{table*}

Table~\ref{tab:tb2_5_5_jobs} reports the additional Terminal-Bench 2.0 validation results using GPT-5.5 as the backbone model. Across five validation jobs, LemonHarness covers 445 trials in total and solves 385 trials, reaching an average accuracy of 86.52\%. The job-level accuracy ranges from 85.39\% to 89.89\%, showing relatively stable performance across different runs. Compared with the GPT-5.3-CodeX submission, this validation setting obtains a higher average accuracy and fewer exceptions per job, suggesting that LemonHarness can benefit from a stronger backbone model while maintaining stable runtime behavior. The remaining failures still mainly come from long-chain tasks that require extended compilation, training, or strict intermediate validation.

\begin{table*}[h]
\centering
\caption{\tb{} 2.0 validation results with GPT-5.5 across five jobs.}
\label{tab:tb2_5_5_jobs}
\begin{tabular}{lccccc}
\toprule
Job date & Trials & Solved & Failed & Exception & Accuracy \\
\midrule
2026-04-26 & 89 & 77 & 12 & 3 & 86.52\% \\
2026-05-06 & 89 & 76 & 13 & 8 & 85.39\% \\
2026-05-07 & 89 & 80 & 9 & 5 & 89.89\% \\
2026-05-15 & 89 & 76 & 13 & 4 & 85.39\% \\
2026-05-15 & 89 & 76 & 13 & 3 & 85.39\% \\
\midrule
Average & 89 & 77.0 & 12.0 & 4.6 & 86.52\% \\
\bottomrule
\end{tabular}
\vspace{-0.1in}
\end{table*}

\subsection{Terminal-Bench 2.1 Supplementary Validation}
Terminal-Bench 2.1 introduces task fixes, continuous validation, and ATIF trajectory requirements. As shown in Table~\ref{tab:tb21}, in supplementary validation with GPT-5.5, the latest three jobs reached an average accuracy of 91.76\%, with 6 total exceptions over 267 trials. This indicates that LemonHarness remains effective under the updated evaluation setting.

The difficult task types in this supplementary validation were similar to those in Terminal-Bench 2.0. They were mainly associated with complex builds, training processes, and multi-step validation. The benchmark update fixed part of the task set, but the main bottlenecks in long-horizon execution remained: long execution chains, complex environment dependencies, and strict validation conditions.
\begin{table*}[h]
\centering
\caption{Supplemental \tb{} 2.1 validation results.}
\label{tab:tb21}
\begin{tabular}{lccccc}
\toprule
Job date & Trials & Solved & Failed & Exception & Accuracy \\
\midrule
2026-05-08 & 89 & 84 & 5 & 2 & 94.38\% \\
2026-05-09 & 89 & 82 & 7 & 1 & 92.13\% \\
2026-05-11 & 89 & 79 & 10 & 3 & 88.76\% \\
\midrule
Average & 89 & 81.7 & 7.3 & 2.0 & 91.76\% \\
\bottomrule
\end{tabular}
\vspace{-0.1in}
\end{table*}

\section{Limitations and Future Work}
The current LemonHarness design is more closely aligned with GPT-style tool calling and model behavior. Its generality across other model families has not been fully validated. Different models vary in tool-call formatting, long-context retention, instruction following, and error recovery. The transferability of the same runtime boundary, general rule knowledge, and time-aware execution mechanism to other model families has not yet been validated. Future work should test LemonHarness across more model families and adjust tool constraints, prompt structure, and knowledge retrieval for different model behaviors.

\section{Conclusion}

This paper presents LemonHarness, an integrated execution framework for long-horizon tasks. The system manages model control, tool execution, task knowledge, and execution records within the same runtime boundary. It also uses generalizable rule knowledge to provide reusable execution rules and acceptance criteria, so the model can start from a clearer task direction instead of relying only on trial and error.
A time-aware execution mechanism further lets LemonHarness rebalance its effort as the budget is spent down, lowering the chance that an otherwise solvable task is lost to long waits or redundant checking. Overall, the design targets common long-horizon failures such as state drift and forgotten constraints, enabling the model to carry out multi-step tool use under a more stable environment state.

\end{document}